\documentclass[10pt,twocolumn,letterpaper]{article}

\usepackage{balance} 
\usepackage{dblfloatfix}
\usepackage[pagenumbers]{iccv}      

\usepackage{booktabs}  
\usepackage{tabularx}  
\usepackage{ragged2e}
\usepackage{multicol}
\usepackage{multirow}
\usepackage{booktabs}
\usepackage{makecell}
\usepackage{marvosym}
\usepackage[T1]{fontenc}

\definecolor{iccvblue}{rgb}{0.21,0.49,0.74}
\usepackage[pagebackref,breaklinks,colorlinks,allcolors=iccvblue]{hyperref}

\def\paperID{16} 
\def\confName{ICCV}
\def\confYear{2025}

\title{When and Where do Events Switch in Multi-Event Video Generation?}

\author{Ruotong Liao$^{1,3}$*, Guowen Huang$^2$*, Qing Cheng$^{2,3}$,Thomas Seidl$^{1,3}$, Daniel Cremers$^{2,3}$, Volker Tresp$\textsuperscript{1,3\ \Letter}$\\
$^1$ Ludwig-Maxilians-University of Munich \\ $^2$ Technical University of Munich $^3$ Munich Center for Machine Learning\\
{\tt\footnotesize ruotong.liao@outlook.com}}

\begin{document}
\maketitle
\renewcommand\thefootnote{*}
\footnotetext{The first two authors contribute equally.}
\begin{abstract}
Text-to-video (T2V) generation has surged in response to challenging questions, especially when a long video must depict multiple sequential events with temporal coherence and controllable content.
Existing methods that extend to multi-event generation omit an inspection of the intrinsic factor in event shifting. The paper aims to answer the central question: \textbf{When} and \textbf{Where} multi-event prompts control event transition during T2V generation.
This work introduces \textsc{MEve}, a self-curated prompt suite for evaluating multi-event text-to-video (T2V) generation, and conducts a systematic study of two representative model families, i.e., OpenSora and CogVideoX. Extensive experiments demonstrate the importance of early intervention in denoising steps and block-wise model layers, revealing the essential factor for multi-event video generation and highlighting the possibilities for multi-event conditioning in future models. 

\end{abstract}
\section{Introduction}
\label{sec:intro}

Text-to-video (T2V) generative models have recently made remarkable progress in creating short video clips from textual descriptions based on diffusion models ~\citep{ho2020denoising,blattmann2023stable, bar2024lumiere}. However, most models and benchmarks focus on single-event or single-shot videos, where a single continuous action or scene is described by one prompt. Extending these models to multi-event long video generation remains challenging. Real-world scenarios often involve sequential events or scene changes (e.g., “a man cooks dinner, then sits down to eat”), which require the model to generate distinct but temporally coherent segments within one video. Existing state-of-the-art T2V models\citep{ruiz2022dreambooth, gal2023textualinversion, deng2019arcface, zhou2024storymaker, tewel2024consistory, zhou2024storydiffusion, wu2023tuneavideo} handle multiple sequential prompts naively: feeding a long prompt with multiple events into a single-generation pass often yields mixed-up sequences without identifying temporal order and deriving natural transitions. Some pioneering works~\citep{kara2025shotadapter, wu2025mind,kim2024fifo} have explored multi-shot or multi-prompt video generation, but they face issues like requiring specialized training data or producing weak prompt adherence and unnatural transitions.

\begin{figure}
    \centering
    \includegraphics[width=1\linewidth]{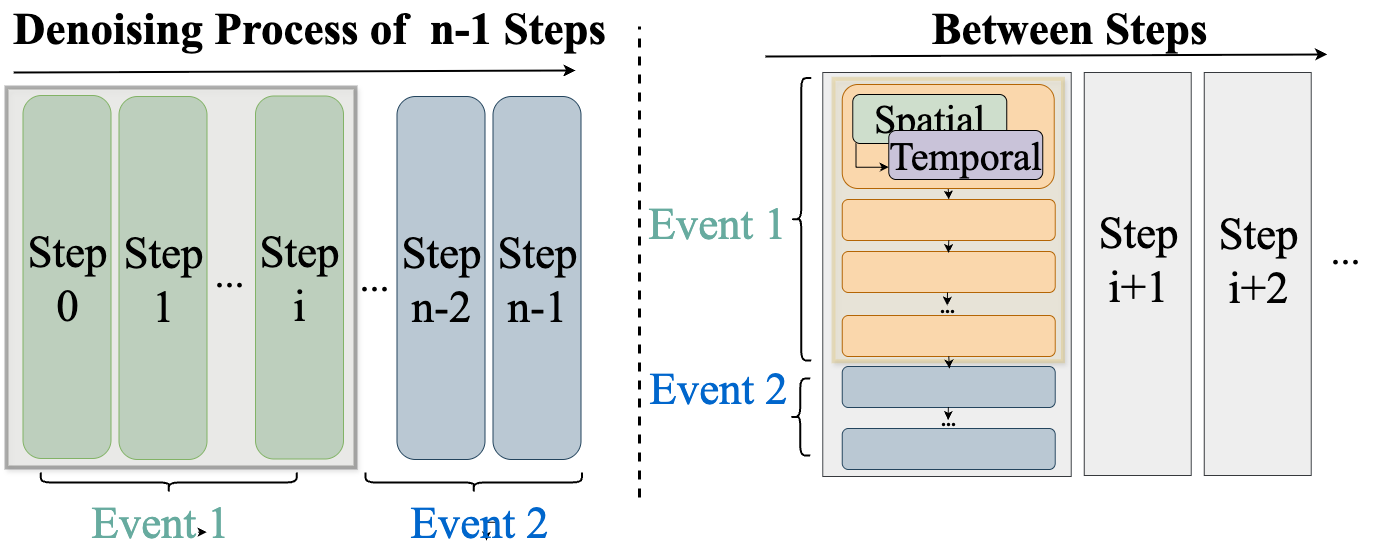}
    \caption{Probing the turning points of multi-event generation}
    \label{fig:illustration}
\end{figure}

In this work, we study how diffusion-based text-to-video (T2V) models can be \emph{probed at inference time} for multi-event generation. We focus on \emph{turning points}, the transition between two events, and probe \emph{when} (denoising steps) and \emph{where} (model layer depth) to inject event prompts. Our findings are: (i) exposing both event prompts within the first $\!30\%$ of denoising steps is dominant for the video content in high-level; (ii) shallow and early blocks govern global semantics, whereas deeper blocks mainly refine appearance and cannot introduce a new event; (iii) late prompt injection is largely occluded by earlier prompt embeddings. These trends hold across model families, with larger video generation models benefiting the most under early steps and shallow layers.
Our contribution can be summarized as follows:
\begin{itemize}
  \item We release \textsc{MEve}, a benchmark prompt suite for multi-event generation, and evaluate two representative families, CogVideo and OpenSora, revealing the dominance of early denoising steps and layers in the intrinsic nature.
  \item We identify two key factors for event switching, i.e., \emph{diffusion denoising step scheduling (when)} and \emph{layer depth (where)}, highlighting multi-event conditioning possibilities for future directions.
\end{itemize}

\section{Benchmarking: \textsc{MEve}}
To evaluate multi–event synthesis, we construct \textsc{MEve}, a prompt suite comprising dual–event descriptions for simplicity from three complementary sources.

\paragraph{LLM as data synthesizer}
Following prior settings~\cite{kara2025shotadapter}, we use Gemini 2.5 Pro ~\cite{gemini} as a controlled prompt generator and author $60$ dual event narratives (e.g., ``$\texttt{event}_1$ then $\texttt{event}_2$''). The prompts target the \emph{general} domain and are written to avoid specific scene priors or content biases.

\paragraph{Prompts of diagnostic content}
To disentangle how subject and action factors shape long-video generation, we adapt VBench~2.0 categories~\cite{zhang2024evaluationagent} into a dual event form. Specifically, we convert prompts from categories \emph{Motion Order Understanding}, \emph{Human Identity}, and \emph{Complex Plot} into two sequential events per prompt while preserving each category's diagnostic intent.

\paragraph{Prompts of viewpoint control}
To isolate the effect of \emph{viewpoint} on event controllability, we construct paired egocentric and exocentric variants of the same underlying event descriptions from video narrations from Ego–Exo4D~\citep{grauman2024egoexo4dunderstandingskilledhuman}. We start by gathering
$50$ single-event prompts from the training set and $50$ from the validation set into dual-event prompts, ensuring that no event overlaps occur within a single prompt. 
For viewpoint control, each prompt is prepended with one of two textual formats:  (a)\textit{Egocentric:} Add the prefix ``Generate a first-person view video...'', where the original subject $C$ is rephrased as ``the camera wearer''. (2) \textit{Third-person:} Add the prefix ``Generate a third-person view video...'', denoting an external viewpoint, where the original subject $C$ is transformed into ``the person''.  
This process results in $50$ pairs of prompts, yielding in total $100$ distinct prompts.

\begin{figure}[t]
  \centering

  \begin{minipage}{\linewidth}
    \centering
    \includegraphics[width=\linewidth]{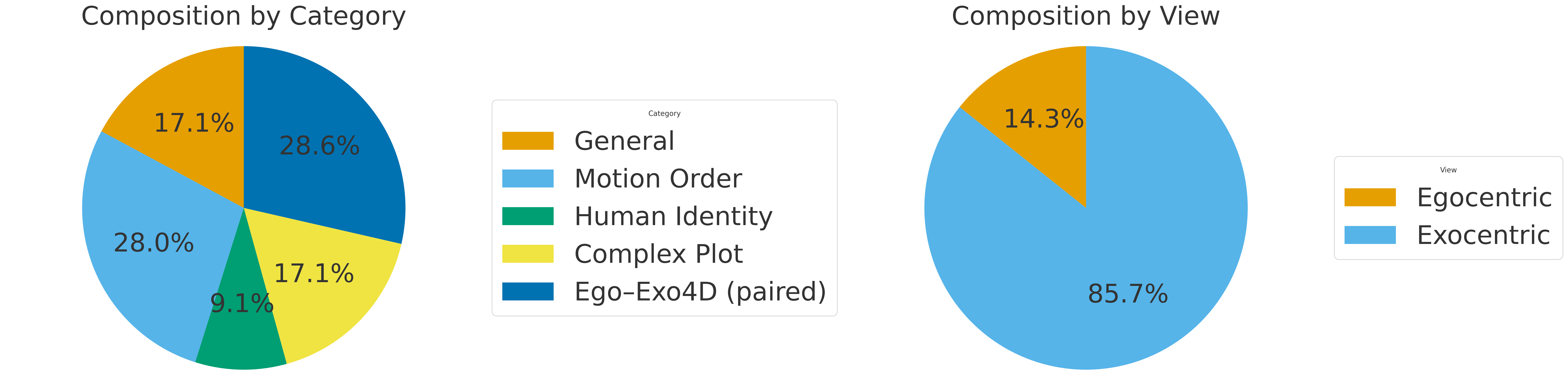}
    \captionof{figure}{Category distribution of the \textsc{MEve dataset.}}
    \label{fig:dist}
  \end{minipage}

  \vspace{0.6em}

  \begin{minipage}{\linewidth}
    \centering
    \small
    \resizebox{\linewidth}{!}{%
      \begin{tabular}{lccc}
        \toprule
        \textbf{Category} & \textbf{View} & \textbf{Events/Prompt} & \textbf{\#Prompts} \\
        \midrule
        General & 3rd & 2 & 60 \\
        Motion Order & 3rd & 2 & 98 \\
        Human Identity & 3rd & 2 & 32 \\
        Complex Plot & 3rd & 2 & 60 \\
        Ego-Exo4D (paired) & 1st \& 3rd & 2 & 100 (50 $\times$ 2) \\
        \bottomrule
      \end{tabular}%
    }
    \captionof{table}{\textsc{MEve} composition statistics.}
    \label{tab:dataset_ratios}
  \end{minipage}

\end{figure}

\section{Methodology: Dynamic Prompt Conditioning}
\label{sec:metho}

We build our approach on diffusion-based T2V generation architectures. For simplicity, we focus on two-event videos in this work, noted by $\text{P}_1$ and $\text{P}_2$, each natural–language clause describing a distinct event.

\paragraph{Can T2V handle multi-events?}
A baseline approach is to simply concatenate $\text{P}_1$  and $\text{P}_2$ into one long sentence.
We concatenate the two event clauses using the connective \texttt{then} to indicate temporal succession as
\[
\mathbf{P} \;=\; [\,\text{P}_1 \;\texttt{then}\; \text{P}_2\,].
\]
$\mathbf{P}$ is then conditioned for the model to generate a full video.
In practice, we found that naive concatenation often leads to the model focusing on the first part of the prompt and either ignoring the second part or blending it incoherently, generating a muddled mix of both events. This motivated more explicit strategies to control when each part of the model influences the generation.

\begin{figure*}[t]
  \centering
  \includegraphics[width=1\textwidth]{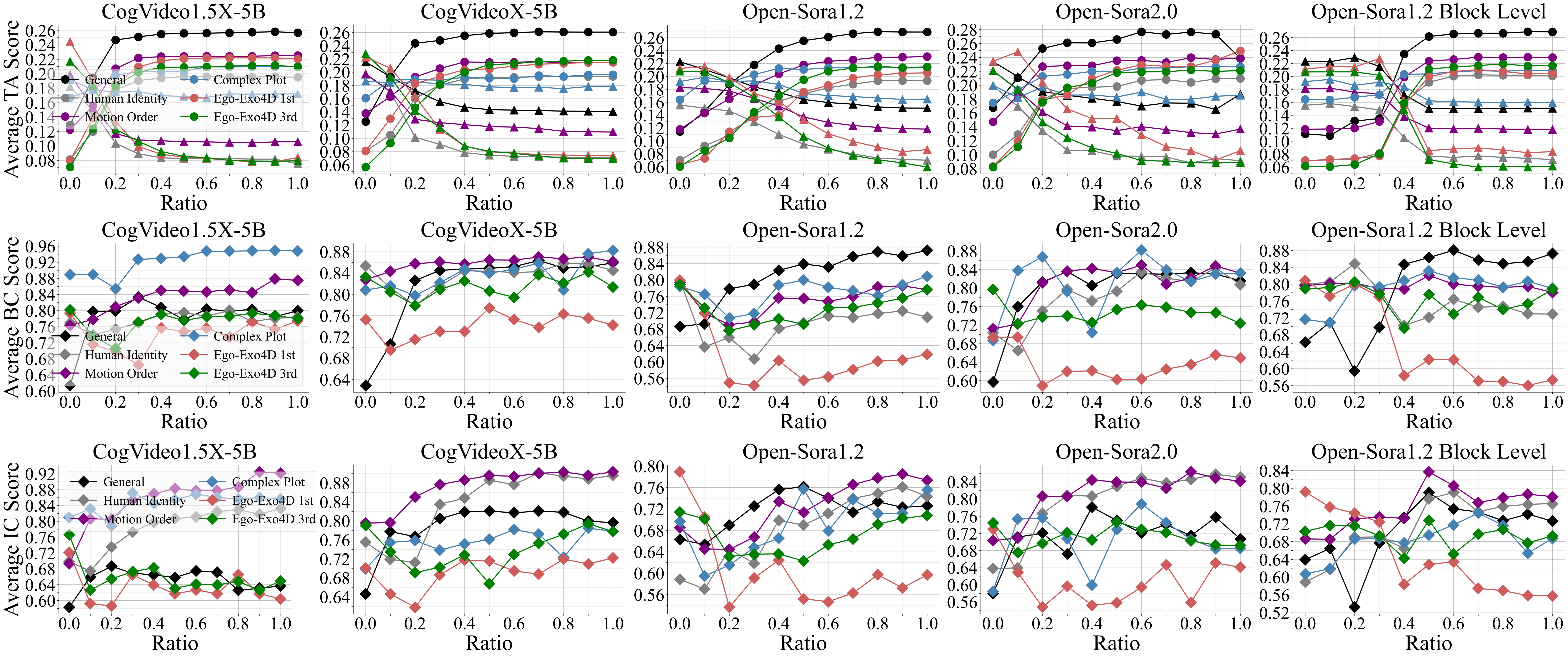}
  \caption{Experiment results on \textsc{MEve} for RQ1 (when) and RQ2 (where), as shown in Columns 1–4 (RQ1) and 5 (RQ2). Curves are category-wise means over prompts; the X-axis is the fusion ratio (RQ1) or the block split ratio (RQ2) $x\in[0,1]$. Top row: Text Alignment (TA$\uparrow$). Middle row: Background Consistency (BC$\uparrow$). Bottom row: Identity Consistency (IC$\uparrow$). In the first row of TA, each color pair represents a group of dual-event generations. Triangle dots denote $\text{P}_1$ related TA, and Round dots denote $\text{P}_2$ related TA. }
  \label{fig:wide}
\end{figure*}


\paragraph{When and where do prompts change events?}
To probe how prompts affect T2V models internally, we study two questions. 
\textbf{(1) RQ1 :} When does the prompt shift events? We investigate \emph{how} the event–prompt shift affects generation dynamics and \emph{when} the temporal \emph{turning point} occurs in denoising steps.
\textbf{(2) RQ2 :} Where does the prompt shift events? We investigate \emph{where} in the DiT hierarchy model layers most strongly influence event realization and the emergence of a turning point. We separate two event prompts $\text{P}_{1}$ and $\text{P}_{2}$ and parameterize the dual–prompt conditioning schedule in the following:

\begin{itemize}
    \item \textbf{RQ1: When do prompts shift events?}

Let $x\in[0,1]$ be the fusion ratio and $\tau\in[0,1]$ the normalized diffusion time.
 and set the conditioning schedule
\[
\mathrm{cond}(\tau)=
\begin{cases}
\text{P}_{1}, & \tau < x,\\
\text{P}_{2}, & \tau \ge x~,
\end{cases}
\]
so that the early diffusion steps, proportion $x$, \textbf{are} conditioned by $\text{P}_{1}$ and the late diffusion steps, proportion $1{-}x$, \textbf{are} conditioned by $\text{P}_{2}$.
For a diffusion process of $N$ steps, the corresponding switch index is $k=\lfloor xN\rfloor$.
For steps $t<k$ we condition on $\text{P}_{1}$; for steps $t\ge k$ we condition on $\text{P}_{2}$.
By sweeping $x$, we probe when the video generation transitions between events and locate the turning point implied by the conditioning.

    \item \textbf{RQ2: Where do prompts shift events within the network?} Analogously, due to model structure constraints, we order the DiT blocks only in OpenSora 1.2 from shallow to deep and let there be $B$ blocks and define $b=\lfloor xB\rfloor$.
At every denoising step, we condition blocks $1{:}b$ on $\text{P}_{1}$ and blocks $b{+}1{:}B$ on $\text{P}_{2}$, thereby allocating an $x$-fraction of shallow blocks to the first event and the remainder to the second.
Varying $x$ reveals how depth–wise allocation conditions the internal shift between events and where the effective turning point is established.
\end{itemize}

Thus, $x{=}1$ yields a $\text{P}_{1}$-only conditioning generation, $x{=}0$ yields a $\text{P}_{2}$-only conditioning generation, and intermediate $x$ places the turning point accordingly.
\section{Experiments}
\label{sec:metho}

\subsection{Experimental Setup}
\paragraph{Models}
We evaluate two representative T2V model families: (1) \textbf{CogVideo}~\citep{hong2022cogvideo}, including CogVideoX-5B~\citep{hong2022cogvideo} and CogVideo1.5X-5B~\citep{yang2024cogvideox}; and (2) \textbf{OpenSora}, including OpenSora 1.2 \citep{opensora}, and OpenSora 2.0 ~\citep{opensora2}. From the denoising step probing, we use the original pretrained CogVideoX-5B, CogVideo-1.5X-5B, OpenSora 1.2, and OpenSora 2.0. For block-wise probing, we only use OpenSora 1.2 as its text embeddings remain independent from the video embeddings during the denoising process across DiT blocks, unlike the others. 

\paragraph{Implementation Details}
The multi-event generation is evaluated on the proposed \textsc{MEve} dataset in the zero-shot setting.
To ensure a fair evaluation, each model generates videos using its native supported frame count and resolution specifications when provided with identical text prompts. For more information, please refer Appendix~\ref{sec:appendix_implementation}. The models are free to allocate frames unevenly according to intrinsic performance. Unless stated otherwise, we sweep $x$ over a fixed grid
$\mathcal{X}=\{0,\,0.1,\,0.2,\,\ldots,\,1.0\} \quad (\text{i.e., } \Delta x=0.1).$

\paragraph{Metrics}
Following commonly used metrics adapted from single–shot T2V protocols~\cite{wu2025mind}, we report three metrics: 
\begin{enumerate}[label=(\alph*), leftmargin=*]
  \item \textbf{Text Alignment (TA)} which assesses the alignment of generated content with text prompts by calculating the similarity between text and video features per event extracted by ViCLIP\citep{wang2023internvid}, averaged across events.
  \item \textbf{Identity Consistency (IC)} which calculates the average DINOv2\citep{oquab2023dinov2}
embedding similarity between the segmented persons using YOLOv5\citep{yolov5}) at the middle frames of each video event;
  \item \textbf{Background Consistency (BC)} measures scene continuity by segmenting background regions with SAM2\citep{kirillov2023segment} and computing DINOv2\citep{oquab2023dinov2} embedding similarity across the midpoint frames of the two events.
\end{enumerate}

\subsection{Experiment Results: Quantative Analysis}

\paragraph{\textbf{RQ1: When to fuse: Early Turning Point}}
As shown in Fig. \ref{fig:wide}, across datasets and models, Text Alignment (TA) increases sharply as the second event prompt becomes visible within the early denoising steps. The turning point comes in $x\in[0,0.3]$, followed by a broad plateau. This indicates that the temporal turning point is set early in diffusion denoising steps: exposing the model to a new event $\text{P}_2$ within the first $\sim 30\%$ steps is sufficient to trigger a shift, while later steps have a diminishing influence. Identity Consistency (IC) remains stable across $x$, suggesting that timing primarily controls \emph{what} event occurs in general, not \emph{who} performs it. Background Consistency (BC) varies depending on whether the event prompt entails a scene change. Except for that, both IC and BC drop in the egocentric viewpoint, likely due to a distribution mismatch, with egocentric footage underrepresented in the training data.

\paragraph{\textbf{RQ2: Where to fuse: Shallow Blocks Govern Global Event Semantics}}
As shown in Fig.\ref{fig:wide}, Block-wise fusion reveals that conditioning to the shallow $xB$ blocks determines video content at a high-level up to $x\approx 0.3$, after which gains saturate. Assigning $\text{P}_2$ to early blocks (with $\text{P}_1$ retained elsewhere, improves TA for the second event and clarifies transitions, while prompts conditioned to mid or late blocks have a limited effect.) The effective turning point emerges when the prompt modulates the early DiT layers, aligning with the view that these layers dominate the coarse spatio-temporal latent representations.
This pinpoints \emph{where} prompts act most effectively: shallow blocks encode the story-level layout and event switch, whereas deeper blocks predominantly refine appearance and content details, and cannot introduce a new event on their own.

\begin{figure}
    \centering
    \includegraphics[width=1.0\linewidth]{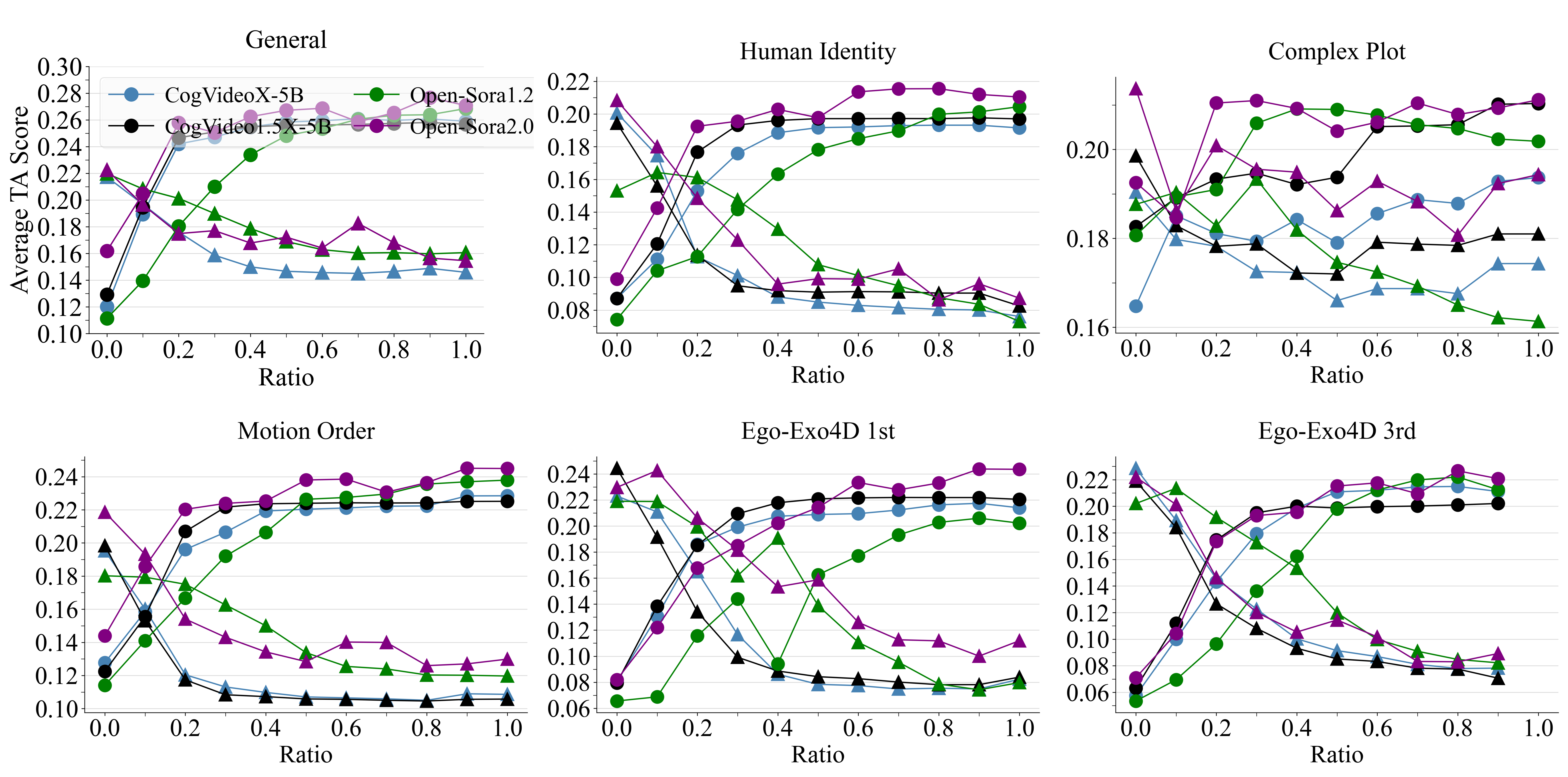}
    \caption{Comparison of models on the same set of prompts per category \ref{tab:dataset_ratios}.}
    \label{fig:modelcompare}
\end{figure}

\paragraph{\textbf{Model capacity modulates, but does not overturn, early-step and shallow-block dominance}}
Figure \ref{fig:modelcompare} shows that more advanced models with larger training capacity show higher TA difference between events, except for \emph{Complex Plot} that shows the expressivity upper limit of the model. But the central finding—that early prompt condition in denoising steps and shallow layer depth is decisive, which holds across model architectures.

\subsection{Experiment Results: Qualitative Analysis}

\begin{figure}[t]
  \centering
  \begin{subfigure}{\linewidth}
    \includegraphics[width=\linewidth]{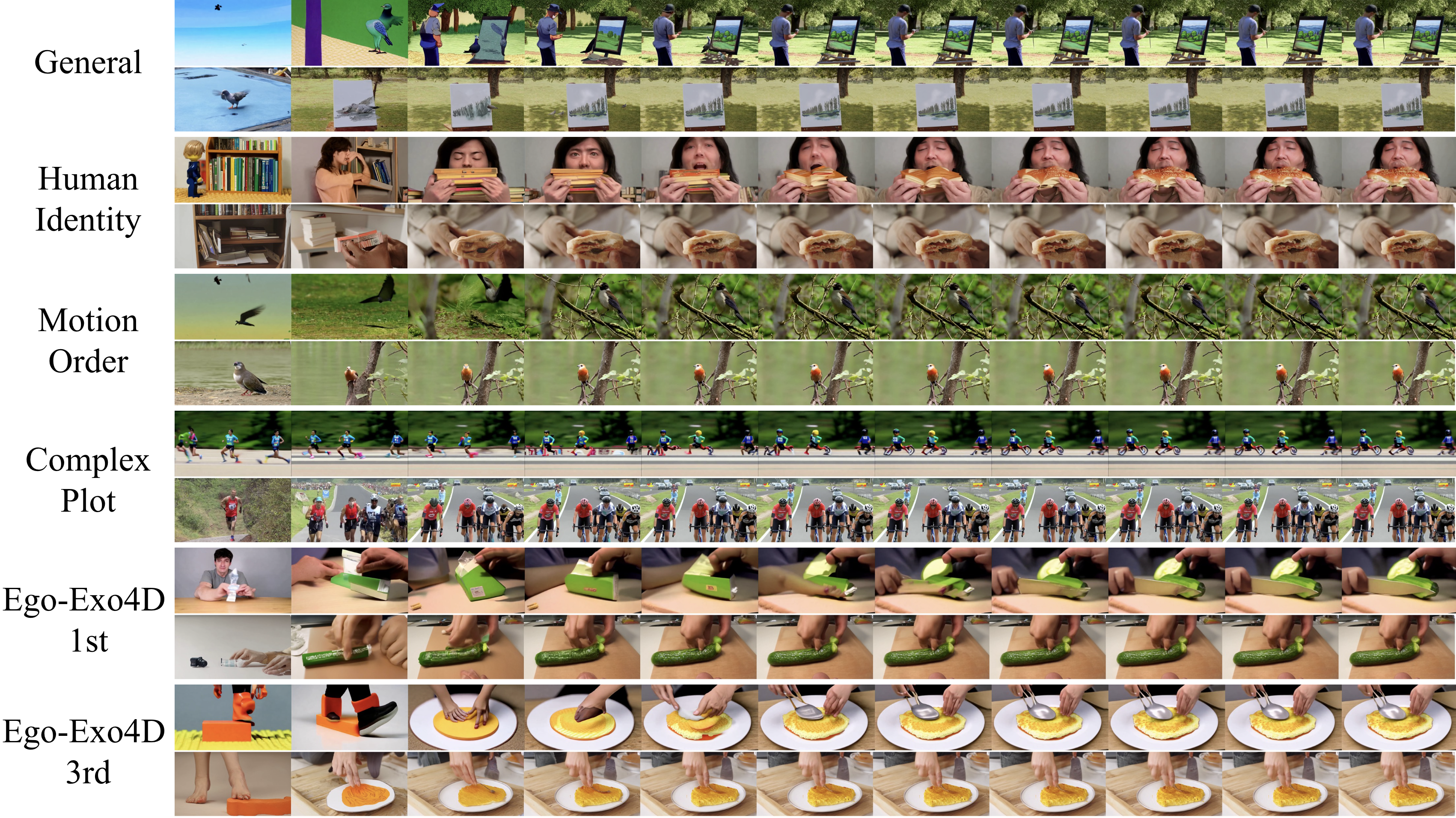}
    \caption{Visualization of the first frame of each video in each category along the ratio axis. Side-by-side grouped comparison between Cogvideo1.5X-5B and OpenSora1.2 conditioned on the same prompt.}
    \label{fig:qa-cogvideo}
  \end{subfigure}

  \vspace{0.5em}
  
  \begin{subfigure}{\linewidth}
    \includegraphics[width=\linewidth]{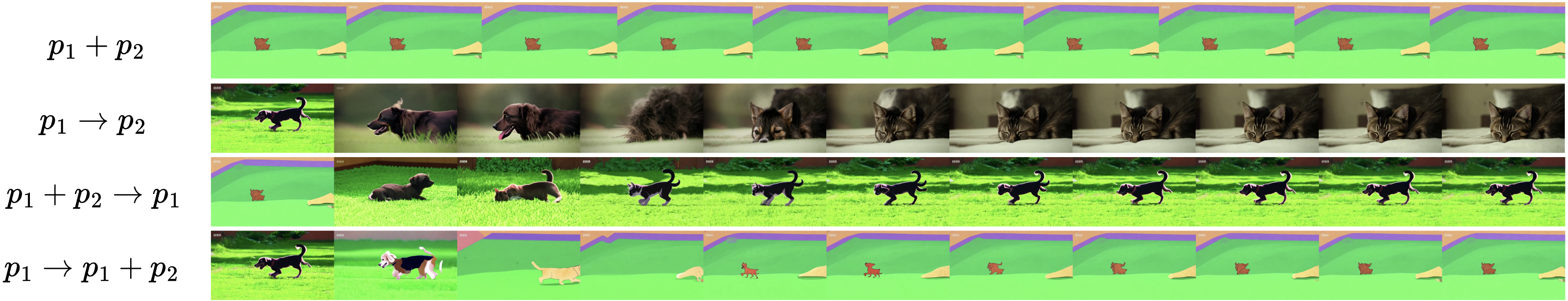}
    \caption{Visualization on the first frame of a video along the ratio axis}
    \label{fig:qa-opensora2}
  \end{subfigure}

  \caption{Qualitative visualizations.}
  \label{fig:qa-stacked}
\end{figure}

\paragraph{\textbf{Naive concatenation underperforms in event shifting}}
To investigate how concatenated events compare with single prompt conditions under various ratios. We provide four settings under various turning point $x=0.3$, visualized in Figure \ref{fig:qa-opensora2}: 
(1) $\text{P}_1 + \text{P}_2$; 
(2) $\text{P}_1\!\rightarrow\!\text{P}_2$; 
(3) Concatenated $\text{P}_1 +\text{P}_2\!\rightarrow\!\text{P}_1$; 
(4) $\text{P}_1\!\rightarrow\!\text{P}_1 +\text{P}_2$; 
Concatenated events frequently yield a mixture of two events. The conditioned video favors the early attended event prompt, leading to occlusion of the later prompt $\text{P}_2$ or $\text{P}_1 + \text{P}_2$ in denoising steps. 

\section{Related Work}

\paragraph{Long Video Generation}
Text-to-Image and Text-to-Video generation has surged with increasing attention\citep{li2021colossalai, peebles2023dit, opendit2024, chen2023pixartalpha, flux2025kontext, greenberg2025demystifyingflux, ma2024latte,hunyuanvideo2024, stabilityai_vae_ft_mse, chen2024dcae, radford2021clip, raffel2020t5, ju2024miradata}. This paper carefully selects two representative model families, CogVideo\citep{hong2022cogvideo, yang2024cogvideox} and Open-Sora\citep{opensora,opensora2}, which are widely used, open-source diffusion Transformer models with reasonable video duration, spatial resolution, and semantic consistency, as the testbed for temporally coherent synthesis for multi-events. CogVideoX series models compress spatiotemporal information with a 3D causal VAE and employ a unified 3D attention mechanism, spatial and temporal, so that every latent video patch attends across space and time. Open-Sora progresses from factorized spatial and temporal attention to unified token sequences and hybrid spatial-temporal blocks.

\section{Conclusion}
\label{sec:metho}

This paper highlights a “turning point” in the multi-event video generation models. Prompt conditioning in the early denoising steps, and in shallow layers, is dominant for video contents at the high-level in the T2V models, while late denoising steps and diffusion blocks barely affect.

\section*{Acknowledgement}
This work is supported by the Munich Center for Machine Learning. 
The authors gratefully acknowledge the scientific support and HPC resources provided by the Erlangen National High Performance Computing Center (NHR@FAU) of the Friedrich-Alexander-Universit\" at Erlangen-N\"urnberg (FAU) under the BayernKI project. 
BayernKI funding is provided by Bavarian state authorities. Furthermore, we gratefully acknowledge funding from the German Federal Ministry of Research, Technology, and Space under grant 01IS24029 (for the Software Campus project MuKChat).

{
    \small
    \bibliographystyle{ieeenat_fullname}
    \bibliography{main}
}
\appendix

\section {Model Output Specifications}
\label{sec:appendix_implementation}

Due to computational resource limitations during inference, Open-Sora 2.0 was evaluated at a reduced resolution compared to other models. However, this resolution constraint does not fundamentally limit the model's performance, as Open-Sora 2.0 was originally trained on 256px resolution with extensive data, and its 768px capability is derived through upscaling from the base 256px generation process, as detailed in their technical report~\cite{opensora2}. 

We also utilize smaller video frames as single events in the dataset can be adequately captured while simultaneously reducing computational overhead during the experiment.
\begin{table}[h!]
\centering
\label{tab:model_comparison_transposed}
\begin{tabular}{lll}
\toprule
\textbf{Model} & \textbf{\#Frames} & \textbf{Resolution} \\
\midrule
CogVideo1.5X-5B & 81 & 1360x768 \\
CogVideoX-5B    & 49   & 720x480 \\
Open-Sora 1.2   &  96 & 1280x720 \\
Open-Sora 2.0   & 129  & 256x256 \\
\bottomrule
\end{tabular}
\centering
\caption{Inference Details of Video Generation Models}
\end{table}


\section {Dataset Construction}
\label{sec:appendix_prompt}
We employ Gemini 2.5 Pro~\cite{gemini} to generate 60 two-event prompts comprising our General dataset. The model was instructed as follows: 

\texttt{Our experiment involves using text-to-video generation models and the text prompt should contain two events. I would like you to generate a set of prompts, where each prompt contains two events and the two events should be easy to visually distinguish from each other. You should also pay attention to keep the causality of the events.} \\

We additionally leverage Gemini 2.5 Pro~\cite{comanici2025gemini25pushingfrontier} to process the Complex Plot dataset, which contains prompts with complicated narratives. These complex prompts are decomposed into two distinct event parts. Moreover, pronouns such as ``it'', ``he'', and ``she''  in the second event are replaced with their corresponding explicit subjects from the first event to maintain generation quality when the second event constitutes a larger ratio during the experiment.

\end{document}